\documentclass[alpha-refs]{wiley-article}
\usepackage{xcolor, subcaption, float}
\usepackage{ulem}

 \newcommand{\indep}{\perp \! \! \! \perp}

\usepackage{siunitx}

\papertype{Original Article}
\paperfield{Journal Section}

\title{Where the Bee Sucks* -- A Dynamic Bayesian Network Approach to Decision Support for Pollinator Abundance Strategies}

\author[1]{Martine J. Barons}
\author[1]{Aditi Shenvi}
\affil[1]{Department of Statistics, University of Warwick, Coventry CV4 7AL, UK}
\corraddress{M. J. Barons, Department of Statistics, University of Warwick, Coventry CV4 7AL, UK}

\corremail{Martine.Barons@warwick.ac.uk}

\fundinginfo{EPSRC grant  EP/K039628/1, Warwick Food GRP}

\runningauthor{Barons et al.}

\begin{document}

\maketitle

\begin{abstract}
For policymakers wishing to make evidence-based decisions, one of the challenges is how to combine the relevant information and evidence in a coherent and defensible manner in order to formulate and evaluate candidate policies. Policymakers often need to rely on experts with disparate fields of expertise when making policy choices in complex, multi-faceted, dynamic environments such as those dealing with ecosystem services. The pressures affecting the survival and pollination capabilities of honey bees (\textit{Apis mellifera}), wild bees and other pollinators is well-documented, but incomplete. In order to estimate the potential effectiveness of various candidate policies to support pollination services, there is an urgent need to quantify the effect of various combinations of variables on the pollination ecosystem service, utilising available information, models and expert judgement. In this paper, we present a new application of the integrating decision support system methodology for combining inputs from multiple panels of experts to evaluate policies to support an abundant pollinator population. 

% Please include a maximum of seven keywords
\end{abstract}
* \footnotesize{Where The Bee Sucks (There Suck I) This song from The Tempest by William Shakespeare is sung by Ariel, a spirit who is in the service of the sorcerer Prospero. Act Five, Scene 1, Lines 88-95}

%%%%%%%%%%%%%%%%%%%%%%%%%%%%%%%
%%%%%%% Introduction %%%%%%
%%%%%%%%%%%%%%%%%%%%%%%%%%%%%%%

\section{Introduction} \label{sec:intro}

In today’s ever more interconnected world, decision-making in dynamic environments is often extremely difficult despite vast streams of data and huge models within disparate domains of relevant expertise. Decision support can be valuable, but needs to incorporate all the relevant inputs in a clear and coherent way so that a decision making team can make a defensible selection among policy options. In these contexts, such decision centres often need to draw together inferences in dynamic, plural environments and integrate together expert judgements coming from a number of different panels of experts where each panel is supported by their own, sometimes very complex, models. These judgements need to be networked together to provide coherent inference for appropriate decision support in increasingly complex scenarios. A formal statistical methodology to network together diverse supporting probabilistic models needed to achieve this, the integrating decision support (IDSS), was developed in \citep{Smith2016}. Here we capitalise on this exciting new development to construct decision support for policy selection in the domain of pollination ecosystem services.

In 2014 the UK government issued its first pollinator strategy \citep{NPS2014} and more recently the Pollinator Action Plan 2021 to 2024 \citep{NPS2022} and the Healthy Bees Plan 2030 \citep{HBP2020}.  The given reason is that bees and other pollinators are an essential part of our environment and play a crucial role in food production -- they contribute the equivalent of more than £500 million a year to UK agriculture and food production, by improving crop quality and quantity -- and are also vital to our wider, natural ecosystems; critical to our food industry, our green spaces, wider biodiversity and ensuring healthy and productive ecosystems \citep{NPS2022}.

The importance of pollination services to food production in the UK and worldwide are undisputed \citep{Vanbergen2014}. Pollinator-dependent food products are important contributors to healthy human diets and nutrition \citep{Potts2016} and it is estimated that over 70\% of important food crops worldwide are dependent upon pollinators \citep{Klein2007}. Therefore, the status of bees and other pollinators is of significant concern in global food security \citep{Bailes2015, Blaauw2014, Lonsdorf2009, Lucas2017, Ollerton2012, Novais2016}. Pollination has a direct economic value through increasing the yield and quality of insect-dependent crops. In the UK, this includes oilseed rape, orchard fruit, soft fruit and beans. Many agricultural businesses employ migratory bee services in order to ensure adequate pollination of crops \citep{Bishop2016, Gordon2014}. Total loss of pollinators could cost up to \pounds 440m a year, about 13\% of UK income from farming \citep{POST2010}. Insect-dependent crops can be pollinated by hand, but the cost would be prohibitive (estimated at \pounds 1500 million a year), raising the cost of food in the marketplace and increasing food insecurity and nutrition insecurity substantially. 

It is estimated that pollinator loss would reduce world agricultural production by 5\%  overall, reducing the diversity of food available, particularly affecting `Five-a-Day' crops \citep{POST2010}, with the obvious downstream effect of increasing burden of disease and health costs. Of course, humans are not the only beneficiaries of pollination services and the social  and tourist value of the insects themselves, the other wildlife they support and the floral species reliant on them should not be discounted.

The Pollinator Action Plan defines pollinator health as the state of well-being of wild and managed pollinators that allows individuals to live longer and reproduce more, even in the presence of pathogens, and therefore provide ecosystem services more effectively. Pollinator health is a function of pests, parasites, disease, and other anthropogenic stressors, the availability of appropriate nutrition across life-stages, nest-sites, host plants, mating areas, and hibernation sites. Honey bee health also depends on the beekeepers managing them. If pollinator health is high, we would expect a greater abundance of pollinators.

 The UK National Pollinator Strategy \citep{NPS2014} acknowledged that whilst there is an abundance of excellent research in many aspects, the evidence for the system as a whole is patchy. Therefore there is a need for decision support to identify optimal policies across this complex landscape. There is a need to assess population-level impacts of insect pollinator management actions and the link between insect pollinator population size change and drivers \citep{NPS2022, HBP2020}. This paper seeks to contribute to this need using the IDSS paradigm.
 
%%%%%%%%%%%%%%%%%%%%%%%%%%%%%%%%%%%%%%%%%%%%%%%%%%%%%%
%%%%%%% Integrating Decision Support System %%%%%%%%%%
%%%%%%%%%%%%%%%%%%%%%%%%%%%%%%%%%%%%%%%%%%%%%%%%%%%%%%

\section{Integrating Decision Support System} \label{sec:IDSS}

\subsection{Overview} \label{subsec:overview}

An IDSS is a unifying statistical framework that enables comparison of candidate policies within complex and evolving systems. It was introduced in \citet{Smith2016, leonelli2015bayesian}, and has since been successfully applied to a range of applications, including to support decision-making in household food security \citep{barons2020decision}, digital preservation risk \citep{baronssafeguarding}, and for counteracting activities of terrorist groups \citep{shenvi21bayesian}. The IDSS decomposes a complex system into its various distinct components where each component has its own statistical model, overseen by a panel of subject matter experts e.g. weather models overseen by the UK met Office. 

The IDSS framework describes how these component expert models can be combined together using a \textit{composite} model to enable inference and decision-making for the system as a whole. A suitable composite model must satisfy certain sufficient conditions in order to be deemed as \textit{coherent} (see \citet{Smith2016, leonelli2015bayesian} for sufficient conditions leading to a coherent IDSS). Often, the components are determined by existing organisational structures such that each component is modelled by a separate \textit{panel} (e.g. a government department). Each panel can incorporate their domain information and expert judgements -- including any relevant uncertainties -- into their component model. Any available data relevant to the component models is fed through them. Note that the outputs from one component model may be fed into other component models as input. When two or more component models share some internal variables, the most efficient way to ensure consistency, separability and coherence \citep{Smith2016} is to create an additional expert panel for the shared variable, the outputs of which can then feed as inputs into the component models that require that shared variable e.g. if two distinct panels each relied on probability distributions over the future price of oil for energy requirements, it is important that these do not differ, so an additional expert panel estimating this distribution using appropriate data and models and asking the two panels to provide their model estimates conditioned on the oil price panel's estimates ensures the panel separability required for the IDSS paradigm. 

By bringing the various component models together, an IDSS aids decision-makers in evaluating the effects of candidate policies on the various different evolving variables which can influence the outcome variables. These effects are generally quantified into a single score using a well-designed \textit{utility function} \citep{smithbook}, usually taking values in the 0--100 range. Thus, each policy is associated with a utility score designed so higher values signal a more desirable policy in terms of the outcome variables. It is crucial to note here that any uncertainties embedded within the component models are systematically propagated through the composite model and are reflected in the variances associated with each utility score. Higher uncertainties lead to larger variances, and in such cases, the outcome variables are very sensitive to the input values. Further, the utility function can be chosen such that a larger variance leads to lower utility scores (see example 3.15 in \citet{smithbook}). Risk-averse decision-makers, (such as some government bodies), tend to prefer policies that perform consistently over a wide range of input values as compared to those with high associated uncertainties. Therefore, it is important to do a sensitivity analysis with different input values.

The process of developing an IDSS is an iterative one as detailed in \citep{barons2018eliciting}. Therefore, it is recommended that the process begins with the simplest setting in terms of the panels, component models, composite model and utility function, all of which can be refined throughout the process so that they are of the appropriate complexity. The decision-makers also need to agree upon the time granularity that is most natural for the evolution of the IDSS being developed. This decision is influenced by several factors including the granularity of the component models, the data collection regularity, the natural time granularity for the key variables in the utility function and the policy decision-making cycle \citep{barons2020decision}.  The iterative process is continued until the decision-makers are satisfied with the IDSS, thereby deeming it to be ``requisite" \citep{Phillips}.

\subsection{Dynamic Bayesian Networks} \label{subsec:DBN}

Many statistical models satisfy the coherence conditions for an overarching composite model to bring expert model component models together under the IDSS paradigm \citep{leonelli2015bayesian}, including Bayesian networks (BNs), multiregression dynamic models, chain event graphs, Markov networks and influence diagrams.  For our application of assessing candidate strategies for improving the abundance of pollinators, we will use the discrete-time dynamic variant of Bayesian networks known as the \textit{dynamic Bayesian network (DBN)}. 

Bayesian networks are a well-established family of probabilistic graphical models that combine together a statistical model that decomposes a complex system into a collection of conditional independence relationships among its defining variables, and a graph that visually represents these conditional independence relationships. The BN model class was first introduced in \citet{pearl1986fusion} and has been successfully applied to a wide range of domains for reasoning in the presence of uncertainty. In particular, BNs have previously formed the basis of successful risk assessment and decision support tools; for examples in environmental sciences see \citet{phan2019applications, davies2015bayesian, pollino2007parameterisation, johnson2017environmental}. The nodes of a BN represent the variables of interest and a directed edge between two nodes represents informational or causal dependencies between the two variables. These dependencies are quantified by conditional probabilities of a variable given the values assumed by its parent variables in the network. The BN graphical representation can be easily understood without any formal mathematical or statistical training, enabling domain experts and stakeholders to be drawn in more easily into the modelling process. Formally, a BN is defined as follows. 

\begin{definition}[Bayesian Network]
A Bayesian network (BN) $\mathcal{B} = (\mathcal{G}, P)$ is a probabilistic graphical model over a set of variables $\pmb{\mathcal{X}} = \{X_1, X_2, \ldots, X_n\}$. Here $\mathcal{G} = (V(\mathcal{G}), E(\mathcal{G}))$ is a directed acyclic graph (DAG) whose node set $V(\mathcal{G})$ is given by the variables in $\pmb{\mathcal{X}}$, and $P$ is a joint probability distribution over the variables $\pmb{\mathcal{X}}$. The edge set $E(\mathcal{G}) \subseteq V(\mathcal{G}) \times V(\mathcal{G})$ consists of directed arcs such that lack of an edge between two nodes represents conditional independence between the variables represented by the nodes, and similarly, edges between nodes encode conditional dependence. This conditional independence structure allows the joint probability $P$ to be factorised by the chain rule as 
\begin{equation*}
P(\pmb{\mathcal{X}} = \pmb{x}\,|\, \mathcal{G} ) = \prod_{X_i \in \pmb{\mathcal{X}}} P(X_i = x_i | Pa(X_i) = x_{Pa(X_i)})    
\end{equation*}
\noindent where $Pa(X_i)$ is the set of parents of the node $X_i$ in $\mathcal{G}$.
\end{definition}

Thus, each variable in the BN, represented by a node in its graph, has an associated conditional probability table wherein the conditioning variables are its parents in the graph. The graph of the BN encodes the following conditional independence statements
\begin{align}
    X_i \indep Nd(X_i) \backslash Pa(X_i)  \,|\, Pa(X_i) 
\end{align}
\noindent where $Nd(X_i)$ are the non-descendants of $X_i$, i.e. all the variables in $\mathcal{G}$ that do not have a directed path from $X_i$ to themselves. More advanced conditional independence relationships can be read from the graph of a BN using the d-separation theorem \citep{verma1990causal} (also known as the global directed Markov property \citep{lauritzen1996graphical}).

The decomposition through which BNs are constructed make them a natural choice as a composite model within an IDSS. As a composite model, the nodes of the BN represent the panels or component models within the IDSS. The nodes can still be viewed as variables but each node or variable now has its own underlying model. The dependence structure within the BN represents how the panels interact with each other. All the conditional independence statements encoded by a BN can be read using the d-separation theorem. The realism of these statements should then be discussed with the domain experts and decision-makers to ensure that the structural aspects of the BN are requisite before populating it with any numerical estimates. Further, BNs have the advantage that they are fully transparent with how they process data for estimating the conditional probability parameters, and with how the quantitative effects of any new information is propagated through the network to revise these parameter estimates. In fact, data and expert judgement can be combined together to uncover the conditional probability distributions of the variables in the system \citep{o2006uncertain}. Any uncertainties, such as those coming from sampling errors or more often through the subjective probabilistic estimates provided by domain experts, are encoded within the variances of these parameters. Once the structure and parametric estimates of the BN have been agreed upon, it can be used to evaluate candidate policies. The effects of a given candidate policy can be evaluated within the BN by intervening on the variables sought to be changed by the policy, and propagating the effects of these changes through the system. As described in Section \ref{subsec:overview}, the varied effects of any candidate policy are typically quantified using a utility score. In the context of decision-making, this refers to expected utility maximisation (for theoretical details, see \citet{smithbook}). All these factors lend BNs well to transparent and explainable decision-making as well as auditing. Moreover, within a Bayesian framework, they allow for real-time updating which is crucial for decision-support in an evolving situation.

However, one key limitation of using a BN is that it a static model which means that it provides a snapshot of the system at a fixed point in time, called a time-slice. Whilst it can be used in real-time within an evolving system by focusing on decision-making for a single time-slice at each point, it cannot be used for analysing the effects of a policy or decision that occurs over multiple time-slices. Such short to medium term decision making is exactly what is of interest for evaluating pollinator abundance strategies. For this purpose, we instead use a dynamic variant of the BN know as the dynamic Bayesian network (DBN) \citet{dean1989model}. The DBN represents how the system longitudinally evolves in discrete time. 

\begin{definition}[Dynamic Bayesian Network]
A dynamic Bayesian network (DBN) is a dynamic variant of the BN that evolves in discrete time. A DBN, defined over a set of variables $\pmb{\mathcal{X}}(t) = \{X_1(t), X_2(t), \ldots, X_n(t)\}$ representing a time-series, is given by the tuple $(\mathcal{B}_1, \ldots, \mathcal{B}_{n})$ where ${\mathcal{B}_1}$ is the initial BN over $\pmb{\mathcal{X}}(1)$ and each subsequent BN $\mathcal{B}_t$ represents the state of the system at time-slice $t$ over $\pmb{\mathcal{X}}(t)$ for $t \geq 2$. Assuming the system satisfies the first-order Markov property, the BN $\mathcal{B}_t$ is connected to the BN $\mathcal{B}_{t+1}$ by directed inter-slice temporal arcs to represent the effect of the variables at time $t$ on the variables at time $t+1$.
\end{definition}

A common simplification of the DBN is to assume stationarity of the graphical structure and the model parameters over time. Such a DBN is called a 2-time-slice DBN and can be compactly given by the tuple $(\mathcal{B}_1, \mathcal{B}_\rightarrow)$ where $\mathcal{B}_1$ is the initial BN and $\mathcal{B}_\rightarrow$ is the transition BN that describes the dependencies of a variable $X$ at time $t$ given the values of its parents in time-slices $t$ and $t-1$. Edges that begin in one time-slice and end in the next are called temporal edges. The DBN inherits all the advantages of the BN as mentioned above, and thus, is suitable for short to medium term decision-making within an evolving system.

\subsection{Evaluation of Scenarios} \label{subsec:evaluating_scenarios}

Each candidate policy describes a complex scenario. In order for it to be evaluated using a DBN, we require that the candidate policy can be stated in terms of changes to the probabilities associated with one or more of the variables in the DBN. For example, assume that the level of pesticide use is a binary variable in our DBN for evaluating pollinator abundance strategies which is categorised as `High' or `Low'. A policy that provides subsidies for using integrated pest management (IPM) techniques and thereby, reducing the use of pesticides can be stated in terms of changes to the probability that the level of pesticide use is `High'. 

We use the Netica software \citep{netica} provided by Norsys for creating and analysing the DBN. Within Netica, for a node without any parents, it is straightforward to modify the probabilities associated with that variable. The effect of such changes are automatically propagated to the descendants of this node by the software. For nodes with parents, changing the probabilities within Netica results in the effects of these changes also being propagated to its parents. Realistically, a policy can only affect downstream variables. Therefore, for a node with parents, a policy affecting that variable is implemented by setting a fixed value for that variable. For example, the effects of the IPM strategy described earlier can be analysed by setting the level of pesticide use being `Low' with the probability of 1. 

Both of the above methods for evaluating scenarios can be described mathematically using the do-operator for causal inference introduced by \citet{pearl1995causal}. For a variable $X$ whose associated node has no parents in the DBN, its effect on a variable $Y$ is given as
\begin{align}
    P(Y | do(P(X = x))),
    \label{eq:do_adjust}
\end{align}
\noindent which implies changes to the probability distribution of $X$. Similarly, if the node associated with $X$ has parents in the DBN, its effect on $Y$ is given by
\begin{align}
    P(Y | do(X = x)).
\end{align}
Thus, we can calculate the changes to the expected utility score as a result of effects brought about by a given candidate policy.

%%%%%%%%%%%%%%%%%%%%%%%%%%%%%%%%%%%%%%%%%%%%%%%%%%%%%%
%%%%%%%% IDSS DBN FOR POLLINATOR STRATEGIES %%%%%%%%%%
%%%%%%%%%%%%%%%%%%%%%%%%%%%%%%%%%%%%%%%%%%%%%%%%%%%%%%

\section{An IDSS for Pollinator Strategies} \label{sec:idss_for_pollinators}

\subsection{Building the Model} \label{subsec:building_model}

A DBN was built with the objective of evaluating candidate policies aiming to improve the abundance of pollinators in the UK. In particular, we focus on three categories of pollinators:
\begin{itemize}
    \itemsep0em
    \item Honey bees: This group refers to the single species of \textit{Apis mellifera} kept by hobby beekeepers and commercial bee farmers in hives;
    \item Other bees: This group refers to the over 250 species of bees living in the wild in the UK, including bumble bees and solitary bees;
    \item Other pollinators: This group refers to the around 6000 species of insects involved in pollination of crops or wild plants in the UK, including moths, butterflies and hoverflies.
\end{itemize}

There are various variables that affect the survival and pollination capabilities of pollinators. \citet{NPS2022} identifies that pollinator health is impacted by threats such as pests, parasites, disease, and other anthropogenic stressors as well as access and availability of environmental requirements, such as appropriate nutrition (including larval food plants, nest-sites, mating areas and hibernation sites. Although these variables are well-documented, the associated data required for appropriately quantifying their effects is often incomplete. 

\begin{figure}[h]
    \centering
    \includegraphics[width = \textwidth]{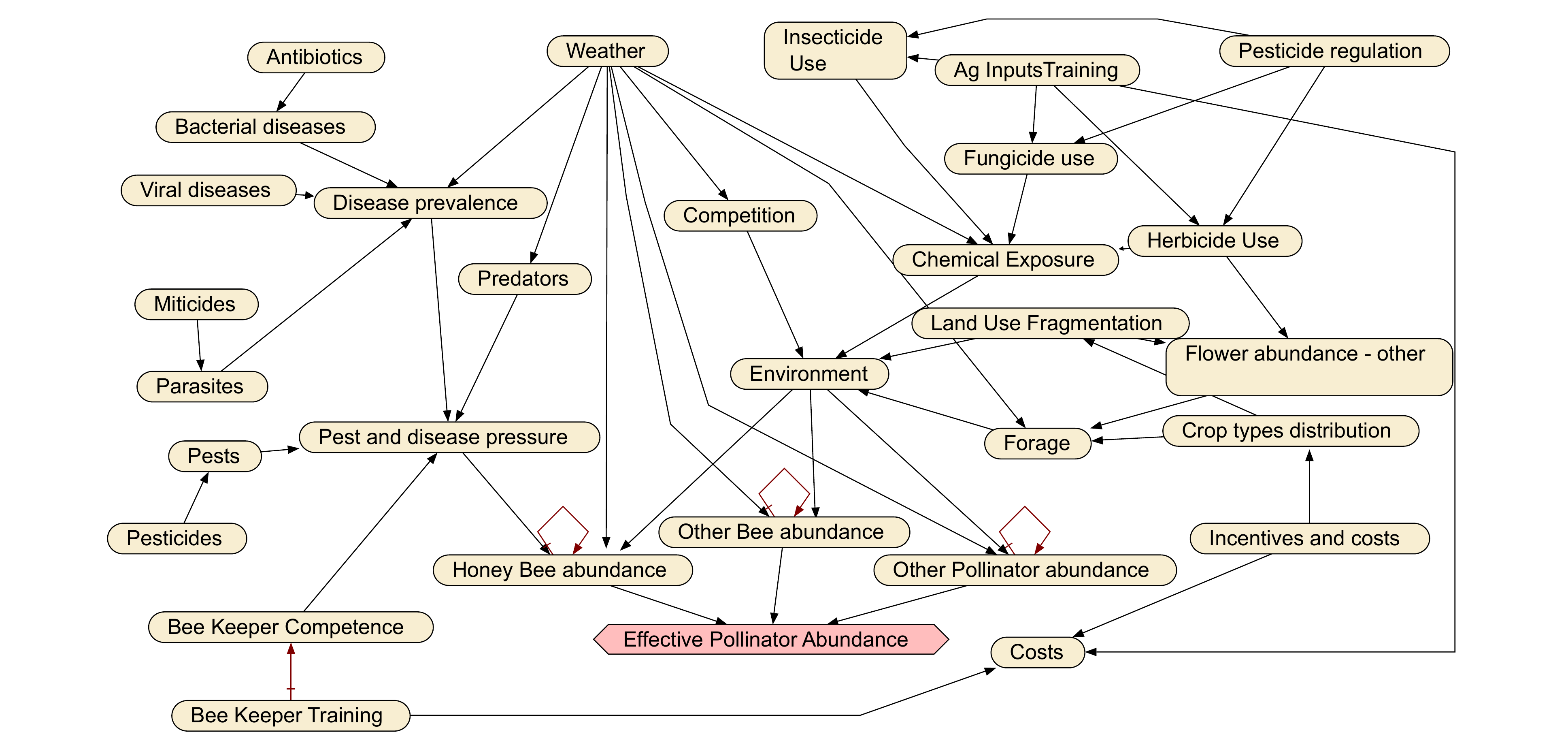}
    \caption{DBN including all variables affecting pollinator populations based on domain literature and expert opinion.}
    \label{fig:full_DBN}
\end{figure}

A literature review was undertaken to identify the variables affecting pollinator populations within the UK context; details of the review can be found in our previous publication, \citep{barons2018assessment}. With the help of pollinator experts, the conditional independence structure between these variables was captured using a DAG such that each variable is included as a node in the DAG. This DAG, shown in Figure \ref{fig:full_DBN}, defines the graphical structure of the DBN that includes the variables affecting pollinator populations. 

The temporal edges in red in Figure \ref{fig:full_DBN} such as the loop for the variable of `Pest and Disease Pressure' and the edge from `Bee Keeper Training' to `Bee Keeper Competence' indicate that the influence of the variable from where the edge emanates has a lag of one time period to affect the variable into which the edge terminates. For example, the loop from `Pest and Disease Pressure' to itself indicates that the level of pest and disease pressure at time $t$ influences the pest and disease pressure at time $t + 1$ for $t \geq 1$. Continuous variables such as weather and pesticide use are discretised since the DBN is a discrete state space model. Further, note that the data available for many of the variables in the DBN in Figure \ref{fig:full_DBN} are incomplete. Given the absence of complete data, probabilities will need to be elicited to populate the conditional probability tables associated with the variables of the DBN. The elicitation exercise is considerably easier and more intuitive when domain experts are asked to elicit probabilities for discrete categories rather than to make judgements about probability distributions \citep{o2006uncertain}. Moreover, if the discretisation is done carefully, the granularity added by retaining variables as continuous does not qualitatively affect the decision-making process.  

\subsection{Expert Panels for the IDSS} \label{subsec:expert_panels}

Having identified the key variables and their conditional independence relationships in Figure \ref{fig:full_DBN}, we then identify the natural panels of experts who might oversee those variables. For example, the Met Office would be best placed to provide information relating to the `Weather' variable, whereas various different units at the Department for Environment, Food and Rural Affairs (DEFRA) would be able to provide information relating to variables such as `Predators', 'Disease Prevalence', `Honey Bee abundance', `Other Bee abundance' and `Other Pollinator abundance'. Below we describe how the variables in the DBN in Figure \ref{fig:full_DBN} are organised into realistic panels based on the UK organisational setting:

\begin{itemize}[itemsep=0pt, topsep=0.5em]
    \item \textbf{Weather}: Weather;
    \item \textbf{Disease and Pest Pressures}: Disease prevalence, Pest and disease pressure, Predators, Parasites, Miticides, Pests, Pesticides, Viral diseases, Bacterial diseases, Antibiotics;
    \item \textbf{Pesticide Use}: Insecticide use, Pesticide regulation, Agricultural inputs training, Fungicide use, Herbicide use, Chemical exposure;
    \item \textbf{Land Use Fragmentation}: Land use fragmentation;
    \item \textbf{Food Supply}: Forage, Crop types distribution, Flower abundance;
    \item \textbf{Social Attitudes \& Incentives}: Incentives; 
    \item \textbf{Environment}: Competition, Environment.
\end{itemize}

\begin{figure}[h]
    \centering
    \includegraphics[width = \textwidth, trim = {0, 0cm, 0 ,0}]{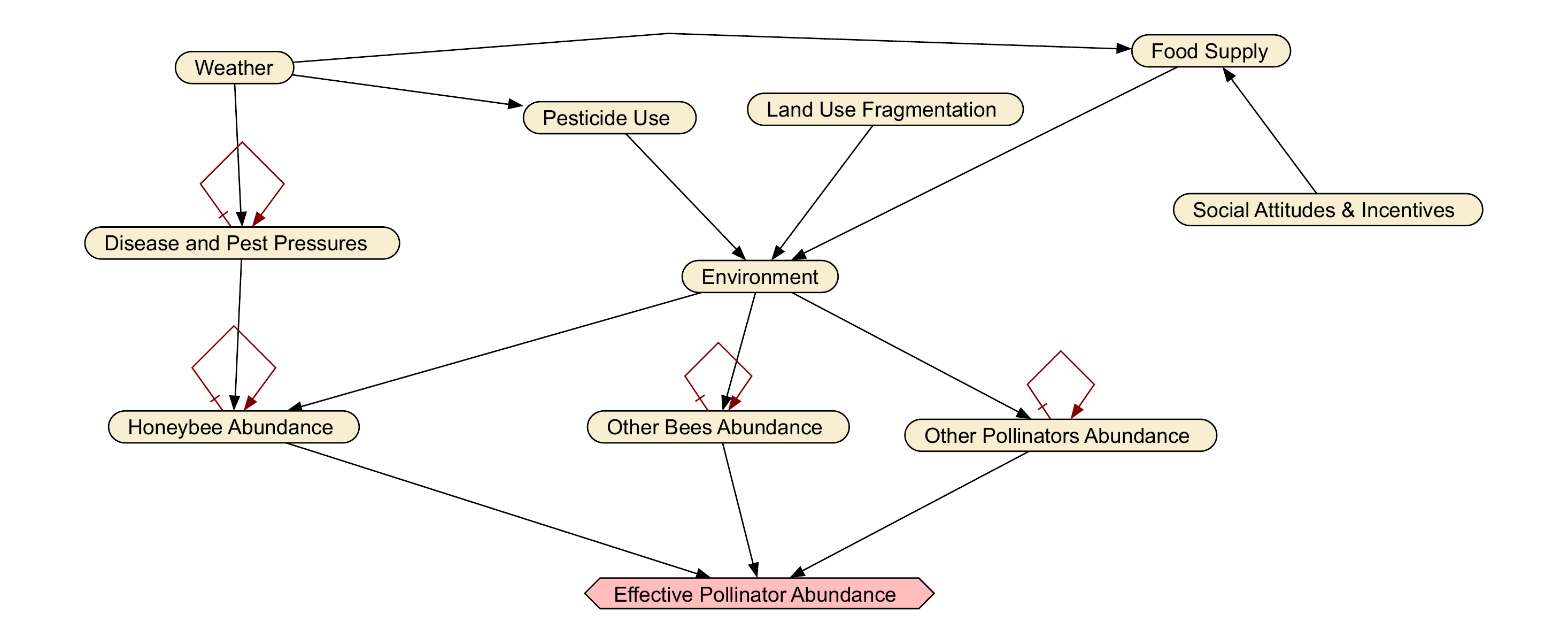}
    \caption{Panel-view of the DBN for evaluating pollinator abundance strategies.}
    \label{fig:DBN_panels}
\end{figure}

The panels for the abundances of honey bees, other bees and other pollinators are kept separate as in the DBN in Figure \ref{fig:full_DBN}. Thus, the DBN in Figure \ref{fig:full_DBN} can be decomposed into a panel-based structure such as the one shown in Figure \ref{fig:DBN_panels}. Each panel will have its own separate model for the inputs they provide. Transforming the full DBN model into a panel-based DBN makes it an IDSS. We have not included the variables associated with bee keeper training or any associated costs as these are likely to be estimated or decided by the decision-makers. 

Our main contribution in this paper is developing and illustrating the use of this proof of concept IDSS for evaluating candidate policies to improve the abundance of pollinators. Therefore, we will henceforth work with the IDSS shown in Figure \ref{fig:DBN_panels}. Note that whilst the `Costs' variable features in the full DBN, we have not included it into the IDSS which we use for illustrative purposes in the remainder of the paper. In practice, a cost-benefit analysis is an essential aspect in the comparison of candidate policies. However, since the IDSS is designed to aid evidence informed decision making by policymakers, and not automated decision making, considerations not included in the IDSS can be considered alongside IDSS outputs (e.g. political acceptability). One time step in our IDSS is a year, since in the UK most pollinators have a break in brood cycle over winter. However, depending on the level of data available, timing of the decision-making cycle and the type of policies considered, decision-makers are at liberty to decide to use more granularity in the IDSS, e.g. a seasonal time step (see further discussion in Section \ref{sec:discussion}). 

\begin{figure}
    \centering
    \includegraphics[width = \textwidth]{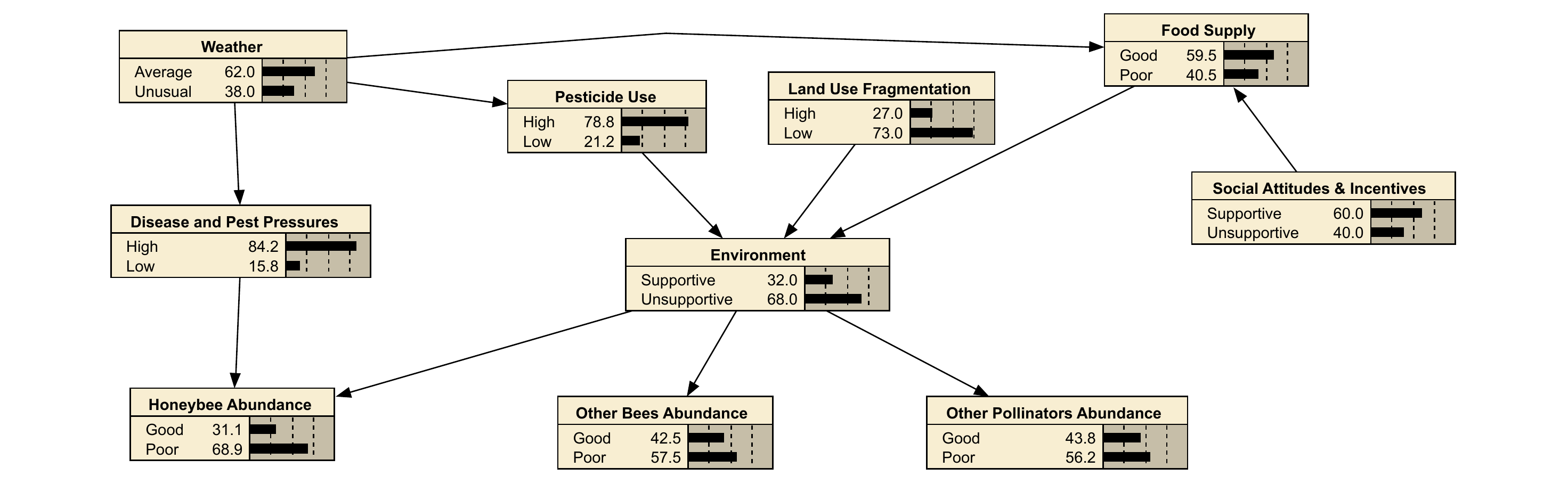}
    \caption{The BN initialising our panel-based DBN IDSS. Note that the parameter estimates for the initialising BN when the DBN is unrolled to 10 time-slices as in Section \ref{sec:scenarios} are not exactly as given here due to the temporal links.}
    \label{fig:compiled_BN}
\end{figure}

In this paper, we directly estimate the outputs of each of the models for the IDSS using a combination of expert elicitation and domain information due to the lack of complete data for all the variables needed as inputs into the various models that form the panel-based DBN. We previously carried out a structured expert elicitation exercise, reported in \citet{barons2018assessment} based on the IDEA protocol \citep{hanea2017nvestigate} to estimate the probability of good pollinator abundance given different weather, disease pressure and environmental scenarios. We use this as a starting point for estimating the parameters in our panel-based DBN. The remaining probabilities are extracted from domain information available in academic journals and technical reports prepared by environmental and governmental bodies (such as the National Pollinator Strategy prepared by DEFRA). Full details of this estimation process are given in Appendix A of the supplementary materials. The probability values for the BN initialising our panel-based DBN is given in Figure \ref{fig:compiled_BN}.

\subsection{Utility} \label{subsec:utility}

To quantify the effect of a given candidate policy, we use the following utility function:
\begin{align}
    \text{Utility} = \dfrac{1}{3} \times p(\text{Honeybees abundance = Good}) + \dfrac{1}{3} \times p(\text{Other bees abundance = Good}) + \dfrac{1}{3} \times p(\text{Other pollinators abundance = Good}).
\end{align}
\noindent This is the simplest form of a linear utility score which gives equal weighting to the abundances of the three types of pollinators that we consider here. The decision makers can adapt this utility function to appropriately reflect their level of risk aversion and priorities. For example, they can choose to use a higher weighting on the abundance of honeybees to reflect the emphasis of abundance strategies typically being on improving the numbers of domesticated honeybees as compared to feral honey bee populations, other wild bees and other pollinators.

\subsection{Sensitivity Analysis} \label{subsec:sensitivity}

We performed a sensitivity analysis to identify the variables that had large effects on the abundances of honey bees, other bees and other pollinators. This also enables us to see how the probabilities for the abundances change under different conditions. Since the panel-based DBN we use is a two time-slice DBN (see Section \ref{subsec:DBN}), we perform the sensitivity analysis on abundances at the second time-slice as the first time-slice is the initialising BN. We use the standard metrics of mutual information (based on entropy) and variance of belief (posterior probabilities) which are directly available within the Netica software \citep{netica}. Mutual information ($I(X; Y)$) between two nodes $X$ and $Y$ is defined as follows
\begin{align}
    I(X; Y) &= H(X) - H(X|Y) \notag \\
    &= \sum_{x} \sum_{y} \log_2 \dfrac{p(x,y)}{p(x).p(y)}
\end{align}
\noindent where $H(X)$ is the marginal entropy of $X$, $H(X|Y)$ is the conditional entropy of $X$ given $Y$ and $I(X;Y)$ is measured in bits. The mutual information metric for $X$ and $Y$ is non-negative (ranges from $[0, H(X)]$), symmetric and indicates how much information $X$ communicates about $Y$. Thus, a higher value of $I(X; Y)$ indicates that changes in $Y$ have a strong impact on $X$. Within Netica, $I(X;Y)$ is also expressed as a percentage of the entropy of $I(X)$. The second metric we use, i.e. the variance of belief $S^2(X;Y)$ metric is calculated as follows
\begin{align}
    S^2(X;Y) = \sum{y} \sum{x} p(x,y) [p(x|y) - p(x)]^2.
\end{align}
\noindent $S^2(X;Y)$ ranges from $[0,1]$ gives the squared values of the expected changes in the beliefs of $X$, taken over all its states, as a result of the information contained in $Y$.

Table \ref{tab:honeybees} shows the top ten nodes that affect the abundances of honeybees at the second time-slice. Similar tables are given in Appendix B for other bees and other pollinators. These tables shows the sensitivity of the population of honey bees, other bees and other pollinators to a finding at other nodes in the first and second time-slices (indicated as [1] and [2] respectively after the variable name) in terms of the mutual information score, percentage of entropy of the specific pollinator's abundance and variance of belief. It is clear from these tables that the environment in the first and second time-slices has a strong influence on the populations of all three categories of pollinators. Further, disease and pest pressure has a strong influence on the abundance of honeybees. These sensitivity checks are also useful as a diagnostic tool so that experts can check whether the influence of the various nodes or panels are as expected. 

\begin{table}[]
\caption{Top ten nodes affecting the abundance of honeybees at the second time-slice.} %This table shows the sensitivity of the honeybee populations to a finding at other nodes in the first and second time-slices (indicated as [1] and [2] respectively after the variable name) in terms of the mutual information score, percentage of entropy of `Honeybee Abundance [2]' and variance of belief.
\label{tab:honeybees}
\begin{threeparttable}
\begin{tabular}{l|ccc}
\headrow
\thead{Variable} & \thead{Mutual Information $(I)$} & \thead{Percentage of Entropy} & \thead{Variance of Belief $(S^2)$}\\
Disease and Pest Pressure [2] & 0.06487   &  10.5    &     0.0140673     \\
Environment [2]     &    0.03101  &    5       &     0.0059849     \\
Honeybee Abundance [1] & 0.02988  &   4.81   &   0.0064502   \\
 Disease and Pest Pressure [1]   &  0.01078    &    1.74     &    0.0021470     \\
Other Bees Abundance[2] & 0.00605  &   0.975  &   0.0011491   \\
Other Pollinators Abundance [2] & 0.00574  &   0.924  &   0.0010831   \\
Food Supply [2]     &    0.00358  &   0.577  &   0.0006341   \\
Weather [2]         &    0.00296  &   0.477  &   0.0005228   \\
Pesticide Use [2]   &    0.00173  &   0.278  &   0.0003256 \\
 Environment [1]       &     0.00136    &    0.219    &    0.0002503     \\
\hline  % Please only put a hline at the end of the table
\end{tabular}
\end{threeparttable}
\end{table}

%%%%%%%%%%%%%%%%%%%%%%%%%%%%%%%%%%%%%%%%%%%%%%%%%%%%%%
%%%%%%%%%%%%%%%%%% SCENARIOS %%%%%%%%%%%%%%%%%%%%%%%%%
%%%%%%%%%%%%%%%%%%%%%%%%%%%%%%%%%%%%%%%%%%%%%%%%%%%%%%

\section{Scenarios} \label{sec:scenarios}

The aim of our IDSS is to model the variables influencing the populations of pollinators in the UK ecosystem sufficiently well to analyse the short and medium term effects of various candidate policies. To demonstrate how such an IDSS can be used in practice, we will analyse various hypothetical but realistic policy and event scenarios as they affect pollinator populations over 10 time-slices, i.e. 10 years. These scenarios are inspired by changes in the influential variables affecting pollinator populations (in particular, environment, and disease and pest pressure; see Section \ref{subsec:sensitivity}) that are discussed in key strategic publications such as the National Pollination Strategy \citep{NPS2014, NPS2022}. The first two scenarios focus on policies that can lead to making the physical environment more supportive for pollinating insects, the third focuses on reducing disease and pest pressure on pollinating insects, the fourth combines both of the above and finally, the fifth does not consider a policy but looks at likely impacts of climate change on pollinator populations in the absence of any ameliorating strategies. Each of the five scenarios is described below in detail; the changes in the probabilities associated with `Good' abundances of the pollinators and the utility scores can be seen in Figures \ref{fig:scenarios} and \ref{fig:stacked}. Note that the points on the graphs in Figure \ref{fig:scenarios} are jittered to enable data points under different scenarios that obtain the same value to be visible next to each other. The exact values of the utilities are given in the supplementary material in Appendix C. 

\subsection{Scenario 1: Decreasing Pesticide Use} \label{subsec:scenario1}

There are three expert panels providing estimates to the environment panel as can be seen in Figure \ref{fig:DBN_panels}, namely, pesticide use, land use fragmentation and food supply which is directly affected by social attitudes and incentives. The first scenario we consider is one where we attempt to make the environment more supportive by decreasing the use of pesticides. Reducing the dependence on pesticides is a critical way forward towards making the environment safer for pollinators as pesticides can have severe and adverse effects on bees \citep{henry2014pesticide}. In the UK, pesticides legislation is regulated by the Chemicals Regulation Directorate (CRD), which is a sub-division of the Health and Safety Executive (HSE). The CRD supports the work of the HSE and DEFRA in ensuring pesticides are used safely, without risk to either spray operators, the general public or the environment. Under the National Action Plan for the Sustainable Use of Pesticides, the UK Governments aim to ``reduce the use of pesticides by utilising alternatives and promoting natural processes'' \citep{nap2020}. In line with this, their action plan includes increasing the uptake of integrated pest management (IPM) techniques. Therefore, for our first scenario, we consider the effects of reducing pesticide use through the adoption of techniques such as IPM. 

\begin{figure}
\begin{subfigure}[t]{0.5\textwidth}
    \centering
    \includegraphics[width = \textwidth]{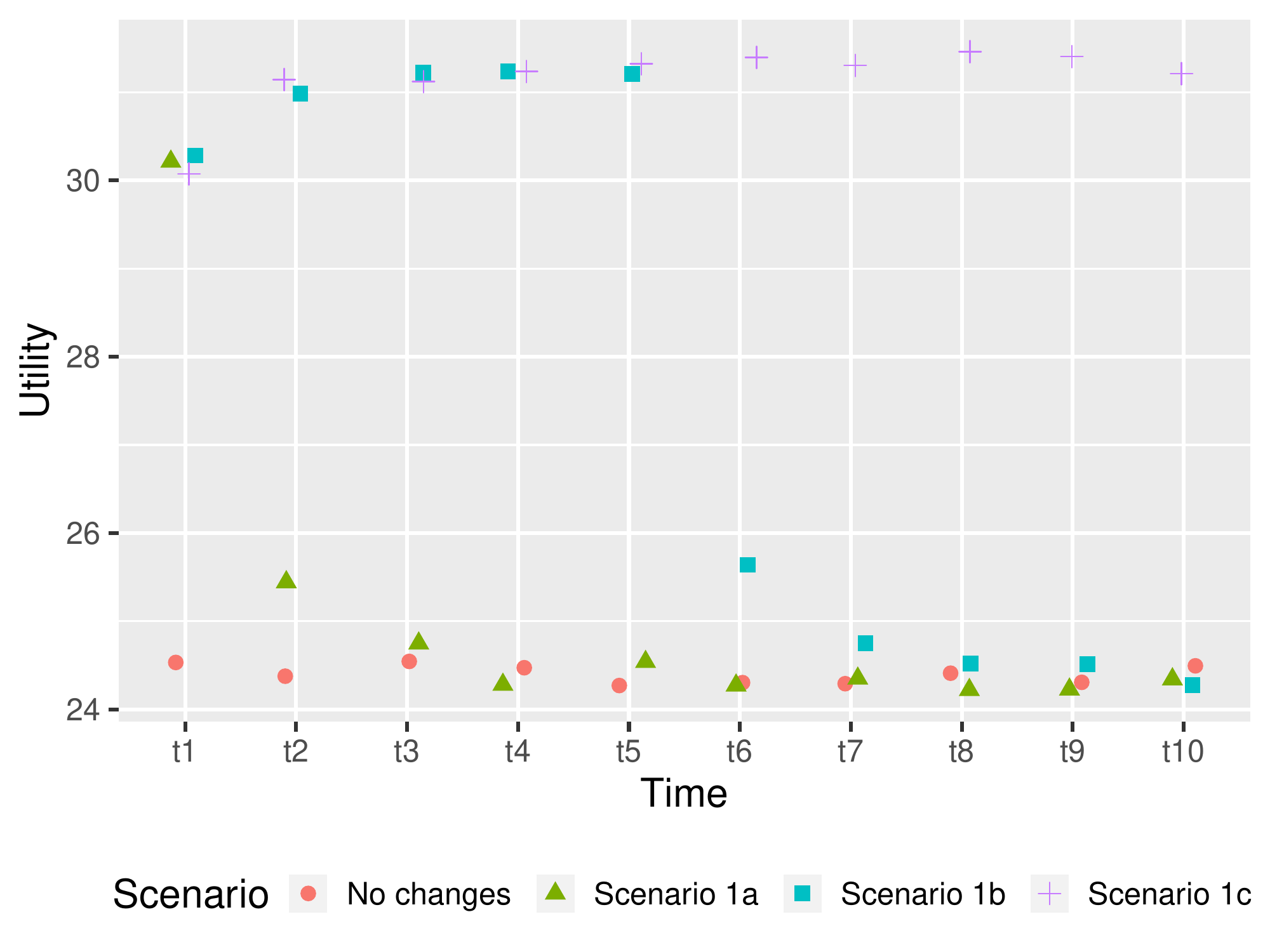}
    \caption{}
\end{subfigure}
\begin{subfigure}[t]{0.5\textwidth}
 \centering
    \includegraphics[width = \textwidth]{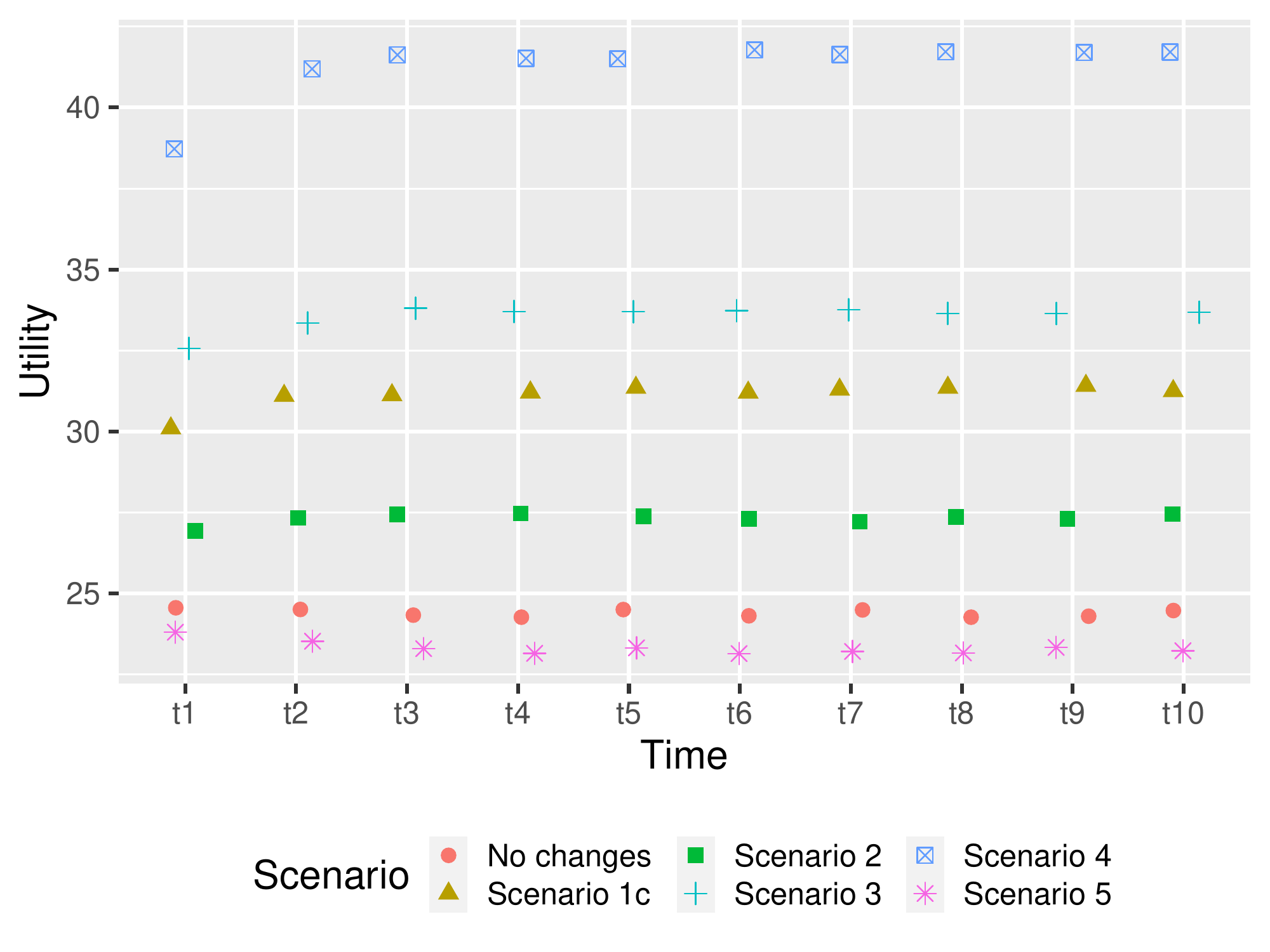}
    \caption{}
\end{subfigure}
\caption{Plots showing the utility scores for each of the ten years under (a) the three cases of Scenario 1 and (b) under Scenario 1c and Scenarios 2, 3, 4 and 5 when compared to having no changes in the IDSS.}
\label{fig:scenarios}
\end{figure}

To analyse the effect of reducing pesticide use, we set pesticide use to ``Low". We consider three cases within this scenario. The first (Scenario 1a) is where pesticide use is set to ``Low" for only one year, the second (Scenario 1b) sets it to ``Low" for five years and the third (Scenario 1c) for ten years. The first two cases assume that once the policy is discontinued, the probabilities revert to pre-policy levels. Under Scenario 1a, we find that there is an immediate effect in the increase in the probability of the environment being ``Supportive" from 32\%\footnote{These probabilities are reported as percentages rather than as values in [0,1] as in our experience, decision-makers find percentages more intuitive.} to 49.3\% and the probabilities for the abundance of honeybees, other bees and other pollinators being ``Good" goes up from 15.8\%, 28.2\% and 29.9\% to 18.6\%, 35.2\% and 36.8\% respectively. Thus, there is a strong positive effect of reducing the dependence of pesticides. However, this effect quickly drops off in the subsequent years. Similarly, under Scenario 1b, the positive effects of the policy tapers off quickly once the policy has been discontinued after five years. Under Scenario 1c, the effects are sustained throughout the ten years. The utility values for each year under the three cases can be seen in Figure \ref{fig:scenarios}(a). 

The drop-off in utility scores when a policy is discontinued can be seen in all other scenarios considered in this paper. In reality, we may expect that there is a lag in observing the effects of a new policy and correspondingly, also a lag in observing a drop off in the effects once the policy has been discontinued. We discuss this further in Section \ref{sec:discussion}. In the remaining scenarios we only analyse the effects of a policy assuming it was implemented over all the years under consideration with no lag in the effects following the implementation of the policy.

% \begin{figure}
% \begin{subfigure}[t]{\textwidth}
%     \centering
%     \includegraphics[width = 0.7\textwidth]{Images/Scenario1_bottom.pdf}
%     \caption{}
% \end{subfigure}
% \vfill
% \begin{subfigure}[t]{\textwidth}
%  \centering
%     \includegraphics[width = 0.7\textwidth]{Images/Scenarios_all_bottom.pdf}
%     \caption{}
% \end{subfigure}
% \caption{Plots showing the utility scores for each of the ten years under (a) the three cases of Scenario 1 and (b) under Scenario 1c and Scenarios 2, 3, 4 and 5 when compared to having no changes in the IDSS.}
% \label{fig:scenarios}
% \end{figure}

\subsection{Scenario 2: Improving Social Attitudes and Reducing Land Use Fragmentation} \label{subsec:scenario2}

For the second scenario, we look at the effects of changing social attitudes and land use fragmentation but not pesticide use on the supportiveness of the environment. In \citet{dicks2016ten}, the authors identify key policies that can be implemented to support pollinators. Among these, they also include policies such as incentives and regulations on various aspects of land use and agricultural inputs, and social marketing and education related to societal and farming support for pollinators. In this scenario, we analyse the effects of changes brought about by policies to improve social attitudes towards measures designed to promote insect pollinator abundance, such as reduced mowing frequency of grass verges in urban areas, and to reduce land use fragmentation, though planning regulations and incentives. 

We set social attitudes to ``Supportive" and land use fragmentation to ``Low". This has the immediate effect of increasing the probability of the environment being ``Supportive" from 32\% to 39.3\%. The probabilities associated with honeybee, other bees and other pollinators populations being ``Good" rise from 15.8\%, 28.2\% and 29.9\% to 17\%, 31.2\% and 32.8\% respectively. These increases are smaller than under Scenario 1; the maximum value of the utility score under Scenario 1c is 31.33 and under Scenario 2 it is 27.33. The utility scores for this scenario over the ten years, given in Figure \ref{fig:scenarios}(b), show that it might be a more effective strategy to lower the pesticide use than to increase social attitudes and reduce land use fragmentation, dependent on consideration of costs. The reasons for this can be seen in the initialising BN in Figure \ref{fig:compiled_BN} which shows that the probability of pesticide use being ``Low" is 21.2\% and has a lot of scope for improvement whereas the probability of land use fragmentation being ``Low" is 73\% which is already quite large. Social attitudes being ``Supportive" is 60\% in the initialised BN and can be improved further but it does not have a direct influence on the environment.

\subsection{Scenario 3: Reducing Disease and Pest Pressure} \label{subsec:scenario3}

As discussed in Section \ref{subsec:sensitivity}, environment is a key variable affecting the populations of all three categories of pollinators, and disease and pest pressure is an additional key variable affecting the honeybee population. As detailed in Appendix A in the supplementary material, we use the level of the parasitic mite \textit{Varroa destructor} as a proxy for disease and pest pressure on honeybees, as our subject matter expert panel advised this this was the most significant hazard to bee health \citep{barons2018assessment} and the direct intervention of beekeepers and bee farmers with miticides, antibiotics, etc. can lead to fast mitigation of this disease burden. UK Beekeepers routinely rely on miticide treatments to keep \textit{Varroa} mites under control \citep{jack2021integrated}. Development of IPM methods to reliably control \textit{Varroa} populations as well as training beekeepers in implementing these methods effectively are important for bringing down disease and pest pressure \citep{jack2021integrated}. Breeding for hygienic behaviour in honey bees is a slower mitigation strategy \citep{Owen2020}. In this scenario, we consider the impact of policies focused on IPM and beekeeper training on the survival and health of the honeybee colonies. 

We consider the effect of setting the disease and pest pressure to ``Low". It has the immediate effect of increasing the probability of honeybee, other bees and other pollinators populations being ``Good" from 15.8\%, 28.2\% and 29.9\% to 39.3\%, 28.3\% and 29.9\% respectively. The effect of this intervention has a very high positive impact on honeybees and a modest impact on other bees and other pollinators compared to the previous two scenarios.  The probability of the managed honeybee population being ``Good" more than doubles under this scenario. Further, this gives us the highest increase in the utility scores of the three policies considered thus far, with a maximum value of 33.77, providing significant motivation for the training of beekeepers in the correct use of miticides.  

\subsection{Scenario 4: Decreasing Pesticide Use, and Reducing Disease and Pest Pressure} \label{subsec:scenario4}

In this scenario, we consider the combination of the two most successful scenarios above, namely Scenario 1c (ongoing reduction in pesticide use) and Scenario 3(reduced pest and disease pressure on insect pollinators). The former positively benefits all three categories of pollinators whilst the latter mostly benefits only honeybee populations. The combination of both these policies gives the highest improvement in the utility score among all scenarios considered in this paper. It has the immediate effect of increasing the probability of honeybee, other bees and other pollinators populations being ``Good" from 15.8\%, 28.2\% and 29.9\% to 44.3\%, 35.3\% and 36.8\% respectively. The utility score reaches and stays at the maximum value of 41.63 from the fourth year. This maximum value is 1.7 times the maximum utility score under no changes to the estimated panel-DBN. 

\subsection{Scenario 5: Worsening Weather Conditions} \label{subsec:scenario5}

In this final scenario, we consider the impact of implementing no new beneficial policies under a warming climate. The UK Climate Projections \citep{ukcp_headline} provides national climate projections for the UK and is developed by the Met Office in collaboration with DEFRA, Department for Business, Energy and Industrial Strategy, and the Environment Agency. The findings report that under a high emission scenario, the seasonal average warming in the UK could amount between 1.3$^{\circ}$C to 5.1$^{\circ}$C in summer and between 0.6$^{\circ}$C to 3.8$^{\circ}$C in winter. By mid-century, hot summers are expected to be 50\% more common. Thereafter, their prevalence strongly depends on whether we are in a low or high emissions scenario. A warming climate adversely impacts the health and population of pollinators, their food supply and their interactions with plants (see e.g. \citet{memmott2007global, dormann2008prediction, hegland2009does, potts2010global}). Under this scenario, we analyse the effects of the weather becoming more unusual. It must be noted that a warming climate is likely to have other far-ranging societal impacts that could have a knock-on effect on pollinators and the various variables affecting them. Here, we only consider the impacts that are quantified within our IDSS.

Since the weather node has no parents in the IDSS (see Figure \ref{fig:DBN_panels}), we can modify the probability distribution associated with this variable as given in Equation \ref{eq:do_adjust} rather than setting it to a specific fixed value. We modify the weather variable to have a 57\% probability of being unusual relative to current data (up from 38\%) and 43\% probability of being average (down from 62\%). The immediate impact of such a change is that the probability of honeybee, other bees and other pollinators populations being ``Good" drops from 15.8\%, 28.2\% and 29.9\% to 14.9\%, 27.5\% and 29.1\% respectively. As expected, the utility scores are the worst under this scenario compared to the earlier four scenarios. By the fourth year, the utility score drops down to 23.3 and then stays there for the remaining six years.

%%%%%%%%%%%%%%%%%%%%%%%%%%%%%%%%%%%%%%%%%%%%%%%%%%%%%%
%%%%%%%%%%%%%%%%%% DISCUSSION %%%%%%%%%%%%%%%%%%%%%%%%
%%%%%%%%%%%%%%%%%%%%%%%%%%%%%%%%%%%%%%%%%%%%%%%%%%%%%%

\section{Discussion} \label{sec:discussion}

We have presented a proof of concept IDSS based on DBNs for comparing candidate policies aimed at improving the population of pollinators. We demonstrated the efficacy of this IDSS in evaluating various scenarios that could be implemented by relevant policy makers. The IDSS presented here consists of variables affecting pollinator populations in the UK based on its unique social, economic and climatic conditions. Therefore, for such an IDSS to be developed and used in another country, steps similar to those detailed in this paper would need to be followed whilst keeping in mind the unique conditions of that country.

A decision-maker can further tailor their specific requirements into this IDSS to fully operationalise it. We used a simple linear utility function for illustrative purposes. A risk averse decision-maker can use a different utility function such as an exponential functional of the form $U(x) = 1 - \exp(-ax)$ for $a > 0$. Due to the uncertainty tower rules, for such a utility function, a high variance in $X$ will be directly influential on the resultant utility score \citep{smithbook}. Further, we note that for illustration, we considered an IDSS that evolves over a yearly time step. In practise, variables such as weather, disease and pest pressure, pesticide use etc. depend heavily on the season. Therefore, a seasonal time step might be considered by policymakers to be more appropriate for this system. This would result in a four time-slice DBN. However, one major drawback to consider here is that in the absence of complete data, this will largely increase the burden of estimating the parameters from domain experts until relevant data can be collected. This has to be weighed against the benefits of using a seasonal time step. The implementation and temporal evolution of the policies under consideration will also need to be considered when choosing the most appropriate time step (see also discussion on choosing the time step in Section \ref{subsec:overview}). 

\begin{figure}
\begin{subfigure}[t]{0.5\textwidth}
    \centering
    \includegraphics[width = \textwidth]{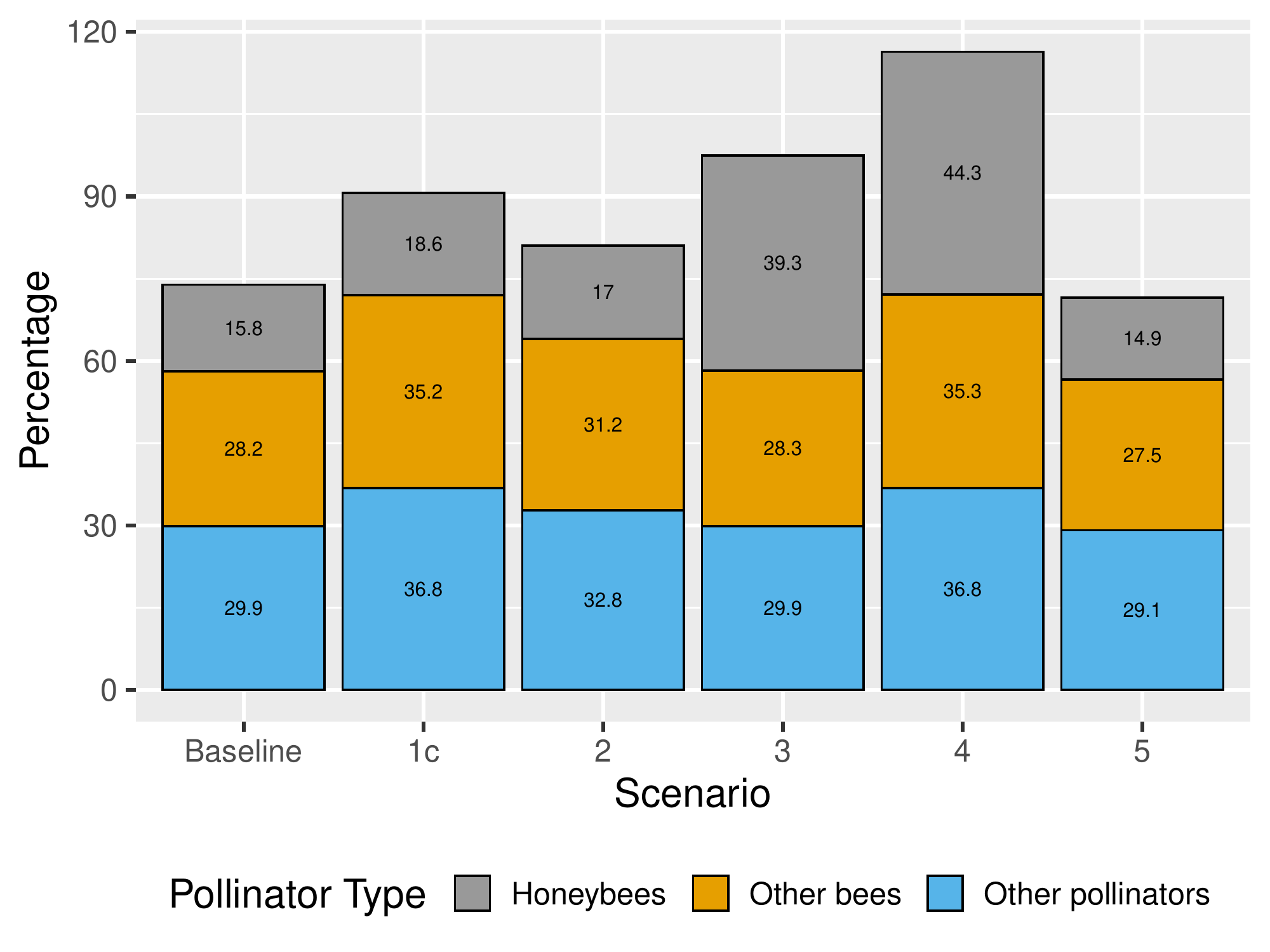}
    \caption{}
\end{subfigure}
\begin{subfigure}[t]{0.5\textwidth}
 \centering
    \includegraphics[width = \textwidth]{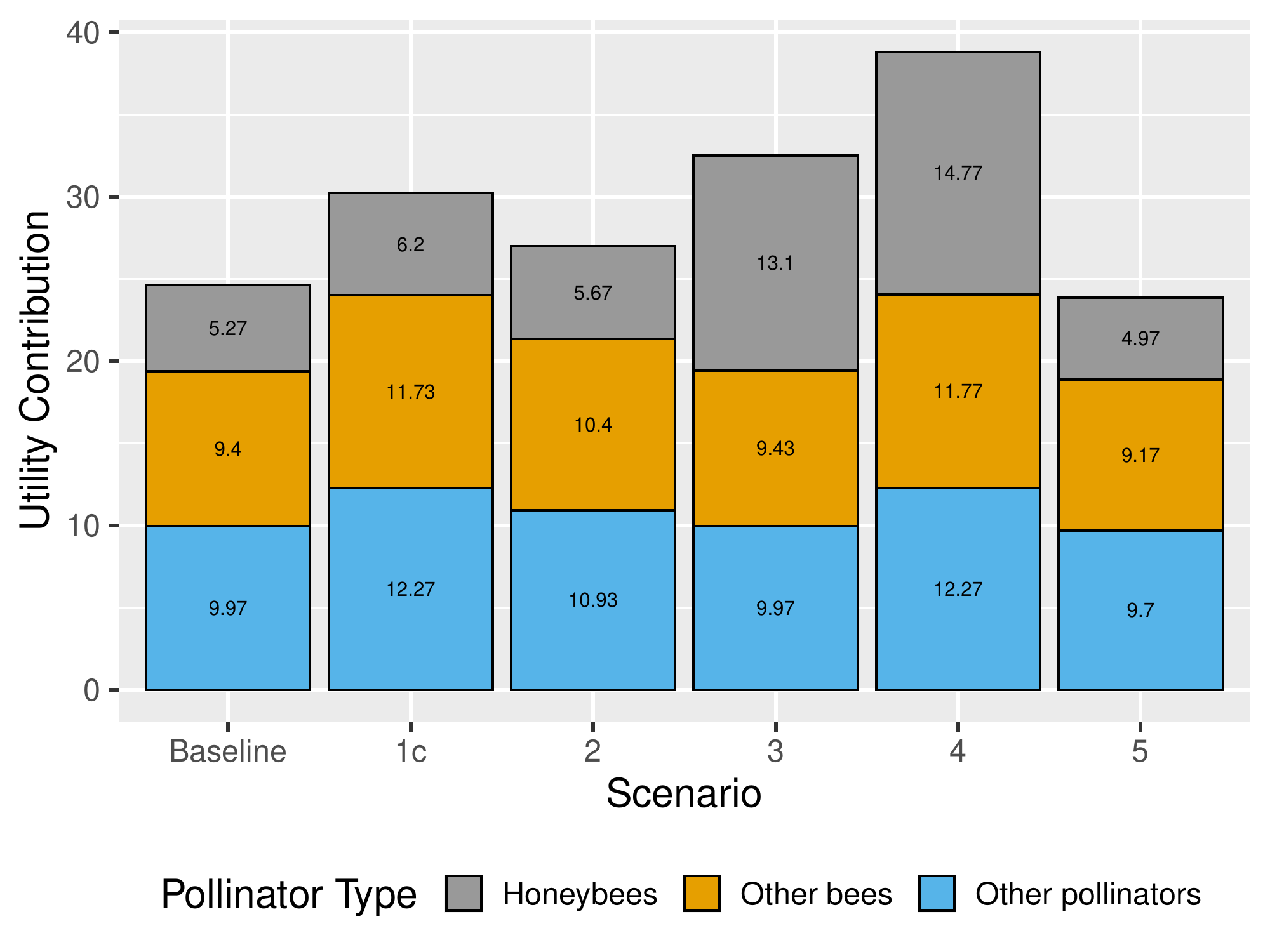}
    \caption{}
\end{subfigure}
\caption{Stacked bar charts illustrating two sets of visualisations of a type that have been used to aid a decision-maker. We have shown in (i) an illustration of the changes in percentages associated with `Good' abundances for the pollinators under each of the five scenarios as compared to the baseline, and (ii) illustration of the changes in contribution to the utility score by the pollinators under each of the five scenarios as compared to the baseline.}
\label{fig:stacked}
\end{figure}

% \begin{figure}
% \begin{subfigure}[t]{\textwidth}
%     \centering
%     \includegraphics[width = 0.6\textwidth]{Images/percentage_bottom.pdf}
%     \caption{}
% \end{subfigure}
% \vfill
% \begin{subfigure}[t]{\textwidth}
%  \centering
%     \includegraphics[width = 0.6\textwidth]{Images/utilitity_contrib_bottom.pdf}
%     \caption{}
% \end{subfigure}
% \caption{Stacked bar charts illustrating two sets of visualisations of a type that have been used to aid a decision-maker. We have shown in (i) an illustration of the changes in percentages associated with `Good' abundances for the pollinators under each of the five scenarios as compared to the baseline, and (ii) illustration of the changes in contribution to the utility score by the pollinators under each of the five scenarios as compared to the baseline.}
% \label{fig:stacked}
% \end{figure}

In Section \ref{sec:scenarios}, the policies we implemented had immediate effects when they were implemented and when they were discontinued. In practise, there will be a lag associated with both types of changes. Due to the immediate implementation assumption, the largest effect of a given policy on the abundances of the pollinators was observed in the same time step that it was implemented for each scenario. Therefore, we found it useful in Section \ref{sec:scenarios} to highlight the changes (as compared to the baseline of `No changes') to the probabilities of the abundances of the pollinators at the first time step for each scenario. It must be noted that good visualisation plays a crucial role in effectively communicating the effects of the various candidate policies to decision-makers; see e.g. \citet{visualisation, climate_risk, baronssafeguarding}. In Figure \ref{fig:stacked}, we illustrate a set of such visualisations for the five scenarios considered in Section \ref{sec:scenarios}. For instance, from Figure \ref{fig:stacked}(a) and (b) it is clear that the beneficial effects of the policy in Scenario 1c is larger for other bees and pollinators than for honeybees, whereas under the policy in Scenario 3, the largest effect is seen for honeybees. Finally, for the decision-makers to choose the most appropriate policy, a cost benefit analysis will be needed to complement the analysis provided by the IDSS.

%%%%%%%%%%%%%%%%%%%%%%%%%%%%%%%%%%%%%%%%%%%%%%%%%%%%%%
%%%%%%%%%%%%%%%%%% CONCLUSION %%%%%%%%%%%%%%%%%%%%%%%%
%%%%%%%%%%%%%%%%%%%%%%%%%%%%%%%%%%%%%%%%%%%%%%%%%%%%%%

\section{Conclusion} \label{sec:conclusion}

In this paper, we have demonstrated that an IDSS based on a DBN can support decision making for policies aimed at improving pollinator abundances. We illustrated, through five different scenarios, how our IDSS can quantify the effects of changes to the system. We have also shown here that for any policy to be beneficial in the long run, it needs to be sustained over that period of time. 

The National Pollinator Strategy identified the key components of the environment and pollinator system which affect insect pollinators. What this work adds is a way to evaluate the efficacy of the policies the National Pollinator Strategy implies in a quantitative manner. This quantification aids policymakers to prioritise actions and combinations of actions that give the greatest probability of good pollinator abundance and allows then to construct a business case based on comparison of cost and benefits.

\section*{Acknowledgements}
Acknowledge with gratitude Robert Owen and Richard Huggins, University of Melbourne, for many helpful discussions.

\newpage 
% Submissions are not required to reflect the precise reference formatting of the journal (use of italics, bold etc.), however it is important that all key elements of each reference are included.
\bibliography{biblio}

\end{document}

\maketitle

\renewcommand{\thesection}{\Alph{section}}

\section{Sources of Parameter Estimation} \label{sec:param_estimation}

In a functioning integrating decision support system (IDSS), expert panels provide estimates for quantities under their purview given each candidate policy under consideration.  Here we draw on published data and estimates in the peer reviewed literature as a realistic proxy within this demonstrator. 

\subparagraph{Panel for `Weather':} Typically the model for this panel would be prepared by the Met Office in the UK context. For our proof of concept illustration, we follow the probability estimates obtained from \citet{barons2018assessment}. They use a decade of Met Office data to create a binary classification of weather as ``Average" (or usual) -- if there were 35 to 70 days in the year with more than 0.2mm of rainfall, total hours of sunshine received was between 240 and 480, and the mean daily temperature fell between 3 and 10$\textdegree$ C, -- and as ``Unusual" otherwise. UK weather was determined to fall in the ``Average" definition approximately 62\% of the time and so ``Unusual" with a probability of 38\%. 

\subparagraph{Panel for `Disease and Pest Pressure':}  \citet{barons2018assessment} states that ``participating experts considered the parasitic mite \textit{Varroa destructor} to be the key pest affecting honey bees". Based on this, we use the \textit{Varroa} prevalence to be a proxy for the disease and pest pressure panel. The UK's National Bee unit, BeeBase, recommends using the conservative threshold of 1000 mites per colony (known as the economic injury level, above which significant harm to bee colonies is likely) to demarcate the level above which the risk posed by the mites is very significant \citep{beebasethreshold}. \citet{kevill2021deformed} reports on the prevalence of two dominant variants of deformed wing virus (DWV), namely DWV-A and DWV-B across the UK. DWV has been shown to be largely driven by the \textit{Varroa} mite acting as a transmission vector\citep{wilfert2016deformed, martin2012global}. In the lack of sufficient data for prevalence of \textit{Varroa} in the UK, we use the prevalence of DWV as a proxy. We categorise disease and pest prevalence as ``High" and ``Low". The probability for ``High" (/``Low") at time $t$ depend on weather and the disease and pest pressure at time $t-1$. The probability of ``High" disease and pest pressure given ``Average" weather is taken to be the average of the prevalence of DWV-A and DWV-B among managed bee colonies, and that given ``Unusual" weather as 10\% higher than that. The probabilities for ``Low" disease and pest pressure are calculated similarly and correspond to prevalences for feral colonies.  Finally, a ``High" (/``Low") disease and pest pressure at time $t-1$ makes it 10\% more likely to have ``High" (/``Low") disease and pest pressure at time $t$.

\subparagraph{Panel for `Pesticide Use':} The level of pesticide use is highly dependent on the weather. ``Unusual" weather may lead to a higher need for pesticides. In particular, the level of use and toxicity (i.e. its effect on pollinators) of pesticides has been found to depend on the temperature \citep{henry2014pesticide}. Detailed data on the level of use of pesticides is difficult to obtain. Instead, we use the level of uptake of integrated pest management (IPM) techniques to be a proxy for level of pesticides used, such that a greater adoption of IPM signals a lower use of pesticides. The probability of ``High" (/``Low") pesticide use is set to be equal to the estimates of the `High IPM' (1 - `High IPM')  uptake levels as given in \citet{creissen2021identifying} under ``Average" weather and 10\% higher under ``Unusual" weather. 

\subparagraph{Panel for `Land Use Fragmentation':} \citet{barons2018assessment} states that the environment is considered to be unsupportive if it has less than 15\% semi-natural-land. Land use change can lead to habitat fragmentation where habitats are destroyed, and those that survive are left smaller, isolated and unconnected. Thus fragmentation due to land use can cause severe and long-term damage to the biodiversity supported by that land \citet{semper2021habitat}. We use the estimate of land fragmentation as ``High" and ``Low" in the UK as given in \citet{land_use}.

\subparagraph{Panel for `Food Supply':} \citet{hicks2016food} states that social factors such as allowing weeds to grow in urban green spaces and public acceptance of planted meadows provide a significant contribution towards maintaining good food supply for pollinators. The estimates within the BN are set using a combination of the effects of social attitudes and weather. 

\subparagraph{Panel for `Social Attitudes':} \citet{schonfelder2017individual} gives the estimates of existence and uptake of beneficial social attitudes for the abundance of bees. We use these to set the probabilities of social attitudes as ``Supportive" and ``Unsupportive".

\subparagraph{Panel for `Environment':} The environment panel takes as input the outputs from the pesticide use, land use fragmentation and food supply panels. Given that each of the panels is estimating a binary variable, for a full elicitation of the environment variable based on its parent variables, we would need to elicit eight parameters. For simplicity, in this proof of concept IDSS, we simplify the estimation to four different scenarios; i.e. estimating whether the environment is supportive (probability of it being unsupportive is obtained as 1 - probability of it being supportive in each case) in the following four cases:
\begin{enumerate}
    \itemsep0em
    \item When all three incoming variables are supportive: probability of ``Supportive" environment is set to 0.8;
    \item When exactly two of the incoming variables are supportive: probability of ``Supportive" environment is set to 0.4;
    \item When exactly one of the incoming variables is supportive: probability of ``Supportive" environment is set to 0.2;
    \item When none of the incoming variables is supportive: probability of ``Supportive" environment is set to 0.05. 
\end{enumerate}

\subparagraph{Panel for `Abundance of Honeybees', `Abundance of Other Bees', and `Abundance of other pollinators':} The effect of the environment and disease prevalence on pollinator abundance is set as given in \citet{barons2018assessment} by marginalising out the weather variable. Note that the effect of disease and pest pressure is higher on honey bees than other bees and other pollinators because \textit{Varroa} was used as a proxy and targets only honey bees. 

\section{Sensitivity of the Abundances of Other Bees and Other Pollinators} \label{sec:abundances}

\begin{table}[h]
\caption{Top ten nodes affecting the abundance of other bees at the second time-slice.} %This table shows the sensitivity of the population of other bees to a finding at other nodes in the first and second time-slices (indicated as [1] and [2] respectively after the variable name) in terms of the mutual information score, percentage of entropy of the `Other Bees Abundance [2]' and variance of belief.}
\label{tab:otherbees}
\begin{threeparttable}
\begin{tabular}{l|ccc}
\headrow
\thead{Variable} & \thead{Mutual Information $(I)$} & \thead{Percentage of Entropy} & \thead{Variance of Belief $(S^2)$}\\
 Environment [2]      &      0.12264    &    14.3      &   0.0356838     \\ 
Other Bees Abundance [1] & 0.02771   &  3.23   &   0.0081066   \\
Other Pollinators Abundance [2] & 0.02219   &  2.59  &    0.0064455   \\
Food Supply [2]   &     0.00947  &   1.11  &    0.0026027   \\
Honeybee Abundance [2] & 0.00605   &  0.706  &   0.0017787   \\
Pesticide Use [2]   &    0.00554  &   0.646  &   0.0016052   \\
 Environment [1]       &     0.00501    &    0.584    &    0.0014274     \\
Land Use Fragmentation [2] & 0.00451   &  0.527  &   0.0012264   \\
Weather  [2]    &         0.00147  &   0.171  &   0.0004077   \\
Other Pollinators Abundance [1] & 0.00091  &   0.106   &  0.0002571   \\
\hline 
\end{tabular}
\end{threeparttable}
\end{table}

\begin{table}[h]
\caption{Top ten nodes affecting the abundance of other pollinators at the second time-slice.} %This table shows the sensitivity of the population of other pollinators to a finding at other nodes in the first and second time-slices (indicated as [1] and [2] respectively after the variable name) in terms of the mutual information score, percentage of entropy of the `Other Pollinators Abundance [2]' and variance of belief.}
\label{tab:otherpollinators}
\begin{threeparttable}
\begin{tabular}{l|ccc}
\headrow
\thead{Variable} & \thead{Mutual Information $(I)$} & \thead{Percentage of Entropy} & \thead{Variance of Belief $(S^2)$}\\
 Environment [2]      &      0.11615     &   13.2     &    0.0348082     \\
Other Pollinators Abundance [1]  & 0.02793  &   3.18   &   0.0083848   \\
Other Bees Abundance [2]  & 0.02219   &  2.53   &   0.0066703   \\
Food Supply [2]      &   0.00891  &   1.01  &    0.0025388   \\
Honeybee Abundance [2] & 0.00574  &   0.653  &   0.0017350   \\
Pesticide Use [2]    &   0.00524  &   0.596   &  0.0015658   \\
 Environment [1]    &     0.00473    &    0.538    &    0.0013923     \\
Land Use Fragmentation [2] &  0.00424  &   0.482  &   0.0011963   \\
Weather [2]       &      0.00138  &   0.157  &   0.0003977   \\
Other Bees Abundance [1] & 0.00091  &   0.103  &   0.0002660   \\
\hline  % Please only put a hline at the end of the table
\end{tabular}
\end{threeparttable}
\end{table}

\section{Utility Scores for the Scenarios} \label{sec:utility}

\begin{table}[h]
\caption{Utility scores for the scenarios in Section 4 over the ten year period.}
\label{tab:honeybees}
\begin{threeparttable}
\begin{tabular}{l|cccccccccc}
\headrow
\thead{Scenario} & \thead{t1} & \thead{t2} & \thead{t3} & \thead{t4} & \thead{t5} & \thead{t6} & \thead{t7} & \thead{t8} & \thead{t9} & \thead{t10}\\
No change & 24.63 &	24.43 &	24.40 &	24.37 &	24.37 &	24.37 &	24.37 &	24.37 &	24.37 &	24.37 \\
1a     &   30.20  &	25.57 &	24.63 &	24.43 &	24.40 &	24.37 &	24.37 &	24.37 &	24.37 &	24.37   \\
1b & 30.20  &	31.13  &	31.27  &	31.33  &	31.33  &	25.7  &	24.67  &	24.43  &	24.40  &	24.37 \\
1c   30.20  &	31.13  &	31.27  &	31.33  &	31.33  &	31.33  &	31.33  &	31.33  &	31.33  &	31.33   \\
2 & 27.00  &	27.30  &	27.33  &	27.33  &	27.33  &	27.33  &	27.33  &	27.33  &	27.33  &	27.33   \\
3 & 32.50 & 33.50 & 33.73 & 33.73 & 33.77 & 33.77 & 33.77 & 33.77 & 33.77 & 33.77  \\
4    &  38.8 & 41.10  & 	41.5 & 41.63 & 41.63 & 41.63 & 41.63 & 41.63 & 41.63 & 41.63  \\
5    &  23.83 & 23.40 & 23.33 & 23.30 & 23.30 & 23.30 & 23.30 & 23.30 & 23.30 & 23.30  \\
\hline  % Please only put a hline at the end of the table
\end{tabular}
\end{threeparttable}
\end{table}

\newpage
\bibliography{biblio}